\documentclass[runningheads]{llncs}
\usepackage{graphicx}
\usepackage{amsmath,amssymb} 
\usepackage{color}

\usepackage[width=122mm,left=12mm,paperwidth=146mm,height=193mm,top=12mm,paperheight=217mm]{geometry}
\begin{document}

\pagestyle{headings}
\mainmatter

\title{Fundamental Matrices from Moving Objects Using Line Motion Barcodes} 

\titlerunning{Fundamental Matrices Motion Barcodes}

\authorrunning{Kasten Ben-Artzi Peleg Werman}

\author{Yoni Kasten ~~~~ Gil Ben-Artzi ~~~~ Shmuel Peleg ~~~~ Michael Werman
}
\institute{School of Computer Science and Engineering \\ 
The Hebrew University of Jerusalem, Israel \\ 
}

\maketitle

\begin{abstract}
Computing the epipolar geometry between cameras with very different viewpoints is often very difficult. The appearance of objects can vary greatly, and it is difficult to find corresponding feature points.
Prior methods searched for corresponding epipolar lines using points on the convex hull of the silhouette of a single moving object. These methods fail when the scene includes multiple moving objects. 

This paper extends previous work to scenes having multiple moving objects by using the "Motion Barcodes", a temporal signature of lines. Corresponding epipolar lines have similar motion barcodes, and candidate pairs of corresponding epipoar lines are found by the similarity of their motion barcodes.

As in previous methods we assume that cameras are relatively stationary and that moving objects have already been extracted using background subtraction.

\end{abstract}

\section{Introduction}

\subsection{Related work}{\label{section:related_work}}
Calibrating a network of cameras is typically carried out by finding  corresponding points between views. Finding such  correspondences often fails when the cameras have very different viewpoints, since  objects and background do not look similar across these views. 
Previous approaches to solve this problem utilized points on convex hull of the silhouette of a moving foreground object.

Sinha and Pollefeys \cite{sinha2010camera} used silhouettes to calibrate a network of cameras, assuming a single moving silhouette in a video. Each RANSAC iteration takes a different frame and samples two pairs of corresponding tangent lines to the convex hull of the silhouette \cite{cipolla2000visual}. The intersection of each pair of lines proposes an epipole.

Ben-Artzi et al. \cite{ben2015camera} proposed an efficient way to accelerate Sinha's method. Generalizing the concept of Motion Barcodes \cite{ben2015event} to lines, they proposed using the best pair of matching tangent lines between the two views from each frame. The quality of line correspondence was determined by the normalized cross correlation of the Motion Barcodes of the lines.   

\begin{figure}
\centering
\includegraphics[width=0.31\linewidth]{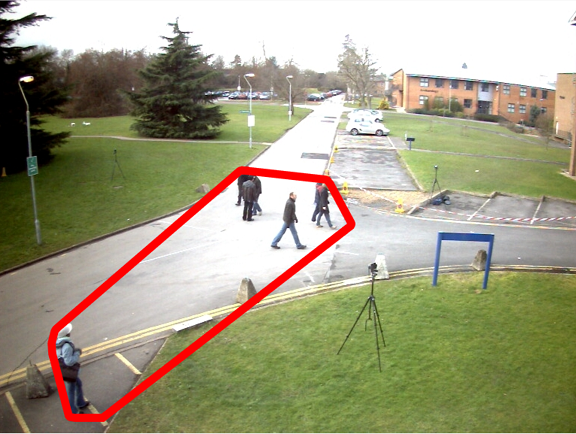} 
\includegraphics[width=0.31\linewidth]{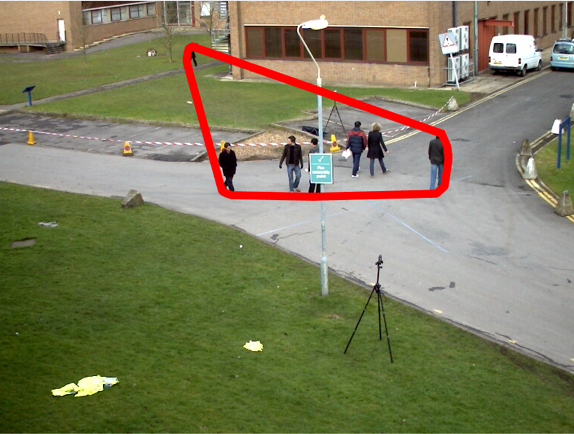}\\
Camera A ~~~~~~~~~~~~~~~~~~~~ Camera B
\caption{Using the convex hull of moving objects fails in scenes with multiple objects. In this case the convex hull ({\it red polygon}) is very different on these two corresponding views, as different objects are visible from the cameras.}
\label{fig:convexHullProb}
\end{figure}

Both methods above \cite{sinha2010camera,ben2015camera} fail when there are multiple moving objects in the scene, as they are based on the convex hull of all the moving objects in the image. In the example shown in Fig~\ref{fig:convexHullProb}, objects that appear only in one of the cameras have a destructive effect on the convex hull. Our current paper presents an approach that does not use the convex hull, and can be used with videos having multiple moving objects.

In other related work, Meingast et al. \cite{meingast2007automatic} computed essential matrices between each pair of cameras from image trajectories of moving objects. They used the image centroids of the objects as corresponding points. However, since for most objects and most different views the centroids do not represent the same 3D point, this computation is error prone. 

Other methods assumed that the objects are moving on a plane \cite{stein1999tracking}, or assume that the objects are people walking on a plane, for both single camera calibration \cite{krahnstoever2005bayesian,lv2006camera} and two camera calibration \cite{chen2007accurate}.

\subsection{Motion Barcodes of Lines}

We address the case of synchronized stationary cameras viewing a scene with  moving objects. Following  background subtraction  \cite{cucchiara2003detecting} we obtain a binary video, where "0"  represents static background and "1"   moving objects.

\begin{figure}
\centering
\includegraphics[height=4mm]{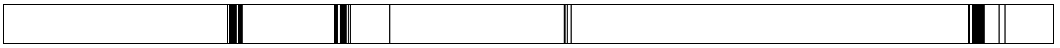}
			
\caption{
A motion barcode $b$ of a line $l$ is a vector in $\{0,1\}^N$.  The value of $b_l(i)$ is "1" when a moving object intersects the line
in frame $i$ ({\it black entries}) and "0" otherwise ({\it white entries}).
\label{fig:point_barcode}}
\end{figure}

Given a video of $N$ binary frames, the Motion Barcode of a given image line $l$ \cite{ben2015camera} is a binary vector $b_l$ in $\{0,1\}^N$. 
$b_l(i)=1$ iff a silhouette, pixel with value 1, of a foreground object is incident to line $l$ in the $i^{th}$ frame. An example of a Motion Barcode is shown in Fig~\ref{fig:point_barcode}.  

\begin{figure}
\centering
	\includegraphics[height=4cm]{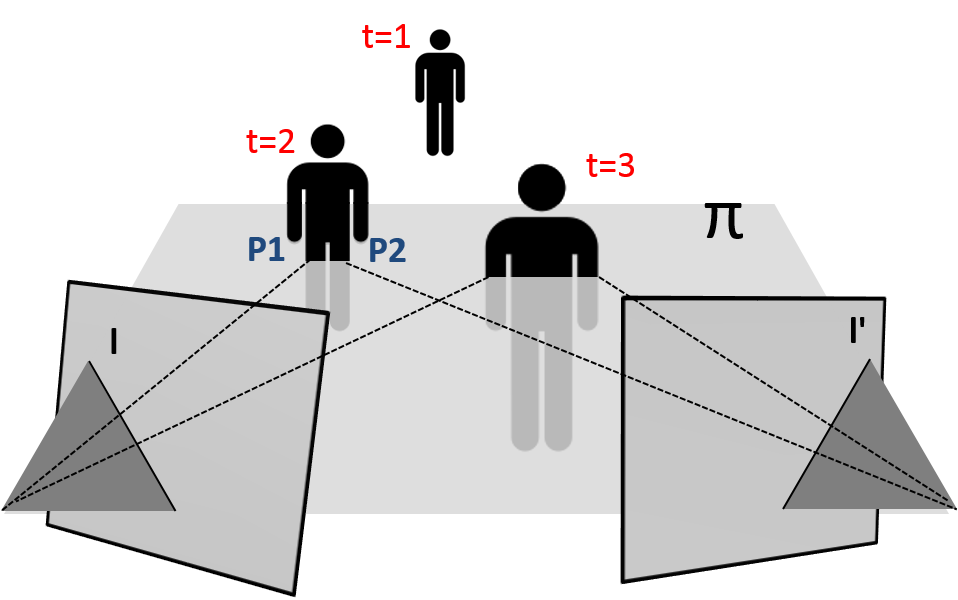}
\caption{Illustration of a scene with a moving object viewed by two video cameras. The lines $l$ and $l'$ are corresponding epipolar lines, and $\pi$ is the 3D epipolar plane that projects to $l$ and $l'$. At time $t=1$ the object does not intersect the plane $\pi$, and thus does not intersect $l$ or $l'$ in the video.  At times $t=2,3$ the object  intersects the plane $\pi$, so the projections of this object on the cameras does intersect the epipolar lines $l$ and $l'$. The motion barcodes of both $l$ and $l'$ is $(0,1,1)$ }
\label{fig:line_barcode}
\end{figure}

The case of a moving object seen by two cameras is illustrated in Fig.~\ref{fig:line_barcode}. If the object intersects the epipolar plane $\pi$ at frame $i$, and does not intersect the plane $\pi$ at frame $j$, both Motion Barcodes of lines $l$ and $l'$ will be $1,0$ at frames $i,j$ respectively. Corresponding epipolar lines therefore have highly correlated Motion Barcodes.

\subsection{Similarity Score Between Two Motion Barcodes}

It was suggested in \cite{ben2015event} that a good similarity measure between motion barcodes $b$ and $b'$ is their normalized cross correlation.

 \begin{align} corr(b,b')=\sum_{i=1}^{N}{\frac{(b(i)-mean(b))\cdot (b'(i)-mean(b'))}{{\lVert b-mean(b) \rVert}_2 {\lVert b'-mean(b')\rVert}_2 }}
 \label{equation:score}
 \end{align}


\subsection{Overall Structure of the Paper}
Our camera calibration approach includes two steps. The first step, described in Sec.~\ref{section:method}, finds candidates for corresponding epipolar lines between two cameras. The second step, Sec.~\ref{section:ransac}, describes how a fundamental matrix between these two cameras is computed from those candidates.
Sec.~\ref{section:experiments} presents our experiments.

\section{Corresponding Epipolar Lines Candidates}
\label{section:method}

Given 2 synchronized videos recorded by a pair of stationary cameras,  $A$ and  $B$, we want to compute their Fundamental Matrix $F$. The Fundamental Matrix $F$ satisfies for each pair of corresponding points, $x \in A$ and $x' \in B$:  

\begin{align} {x'}^TFx=0.
\label{equation:F}
\end{align}

The $F$ matrix maps each point $x \in A$ to an epipolar line $l'=Fx$ in $B$ so that the point $x'$ is on the line $l'$. Any point in image $B$ that lies on the line $l'$ including $x'$ is transformed to a line $l=F^{T}x'$ such that $x$ is on the line $l$. $l$ and $l'$ are corresponding epipolar lines.   $F$ can be computed from points correspondences or from epipolar line correspondences \cite{hartleyMVG}.

In previous methods (\cite{sinha2010camera,ben2015camera}) the convex hull of the silhouette of a moving object was used to search for corresponding epipolar lines.

Our proposed process to find candidates for corresponding epipolar lines does not use the silhouette of a moving object, and can therefore be applied also in cases of multiple moving objects.

Given a video, lines are selected by sampling pairs of points on the  border of the image and connecting them. For each line, the Motion Barcode is computed. We continue only with informative lines, i.e. lines having enough zeros and ones in their motion barcode.

Given two cameras A and B, Motion Barcodes are generated for all selected lines in each camera, resulting $n_1$ vectors in $\{0,1\}^N$  for Camera A, and $n_2$ vectors in $\{0,1\}^N$ for Camera B, where $N$ is the number of frames in the video. 

\begin{figure}
\centering
\includegraphics[width=0.31\linewidth]{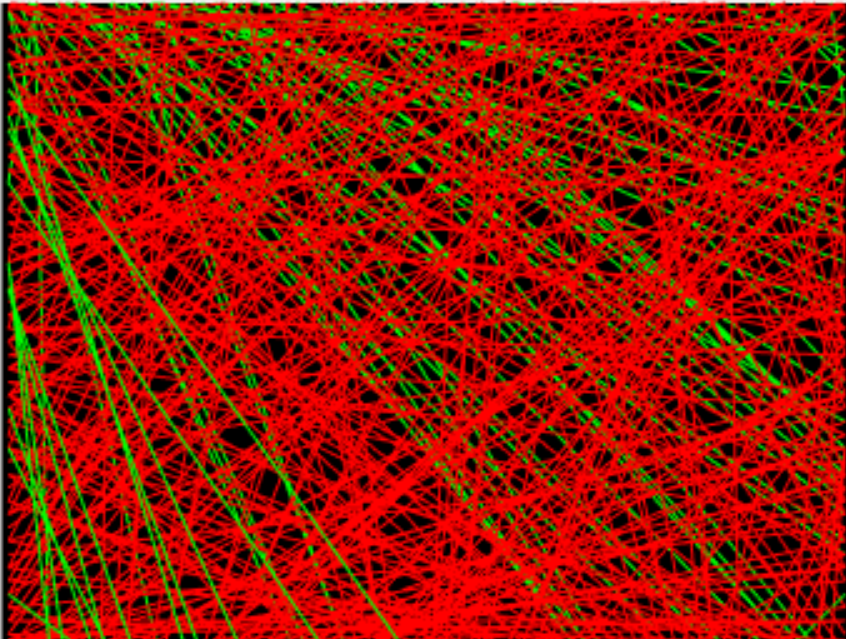}
\includegraphics[width=0.31\linewidth]{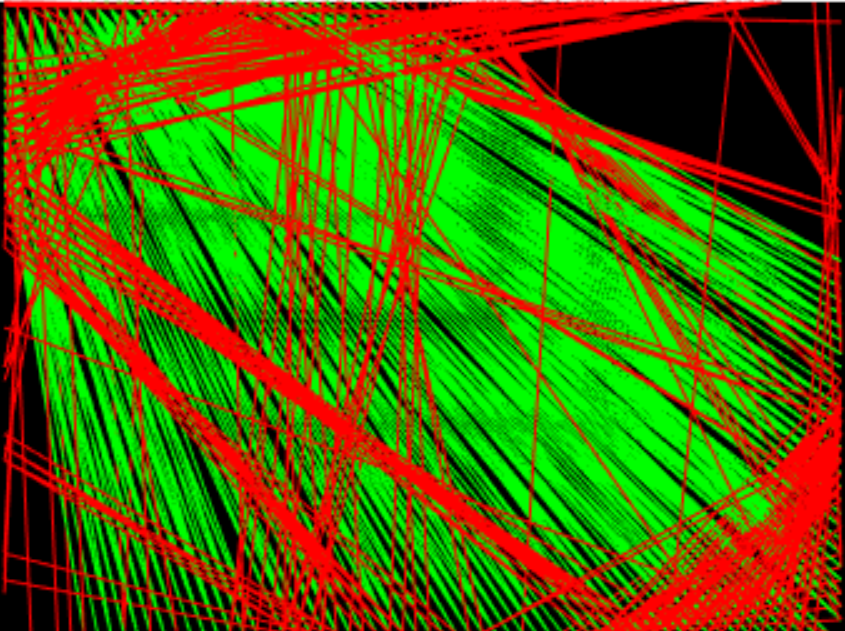}
\includegraphics[width=0.31\linewidth]{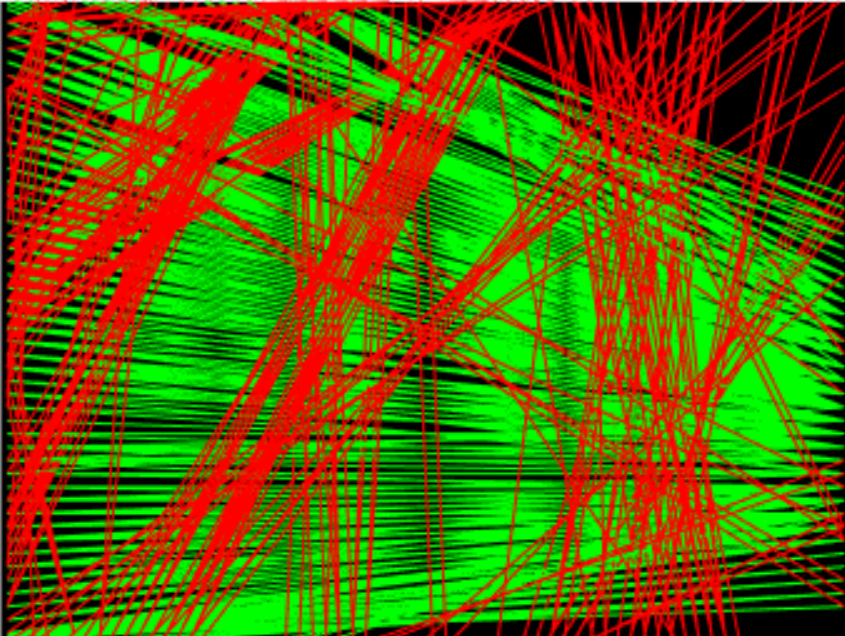}\\
(a) ~~~~~~~~~~~~~~~~~~~~~~~~~~~~~~ (b) ~~~~~~~~~~~~~~~~~~~~~~~~~~~~~~ (c)

\caption{
An example of the effect of filtering candidate pairs using motion barcode similarity. Most datasets are very similar.
(a) Initial candidate lines in Camera A are randomly distributed.
(b) Candidate lines in Camera A having a similar motion barcode to lines in Camera B, as described in Sec.~\ref{section:method}. Most mismatches were removed, and correct epipolar lines dominate.
(c) Same as (b), showing lines in Camera B.
}
\label{fig:correlated-pairs}
\end{figure}

The $n_1 \times n_2$ correlation matrix of the barcodes of the lines selected from Camera A with the lines selected from Camera B is computed using Equation~\ref{equation:score}. The goal is to find  corresponding line pairs from Camera A and Camera B. $1,000$ line pairs are selected using the correlation matrix as follows. For visual results see Fig.~\ref{fig:correlated-pairs}.
\begin{itemize}  
\item If the correlation  of a pair of lines is in the mutual top 3 of each other, i.e. top 3 in both row and column, it is considered  a  candidate.
\item The 1,000  candidate pairs with the highest correlations are taken as corresponding epipolar lines candidates. 

\end{itemize}

\section{Fundamental Matrix from Corresponding Lines}	
\label{section:ransac}

Given a set of candidate corresponding pairs of epipolar lines between cameras $A$ and $B$, our goal is to find the fundamental matrix $F$ between the cameras. 

Experimentally, of the 1000 candidates for corresponding epipolar lines  described in Sec.~\ref{section:method}, about half are correct.  
As not all of our  candidates are real correspondences the algorithm continues using a RANSAC approach.

In each RANSAC trial, two pairs of candidate corresponding epipolar lines are selected. This gives two candidates for epipolar lines in each camera, and the epipole candidate for this camera is the intersection of these two epipolar lines. Next, an additional pair of corresponding epipolar lines is found from lines incident to these epipoles. The homography $H$ between corresponding epipolar lines is computed from these three pairs of epipolar lines. This is described in detail in Sec.~\ref{section:Hcompute}.

The proposed homography $H$ gets a consistency score depending on the number of inliers that $H$ transformed successfully as described in Section~\ref{section:Hconsistency}.

Given the homography $H$, and the epipole $e'$ in $B$, the fundamental matrix $F$ is \cite{hartleyMVG}: 
\begin{align}\label{equation:fFromH}F=[e']_x H^{-T} \end{align}

\subsection{Computing the Epipolar Line Homography}
\label{section:Hcompute}

We compute the Epipolar Line Homography using  RANSAC. We sample pairs of corresponding epipolar line candidates with a probability proportional to the correlation of their Motion Barcodes as in Eq.~\ref{equation:score}.
Given 2 sampled pairs $(l_1, l_1')$ and $(l_2, l_2')$, corresponding epipole candidates are computed by: $e = l_1 \times l_2$ in Camera A, and $e' = l_1' \times l_2'$ in Camera B. 
Given $e$ and $e'$, the distances from these epipoles of each  of the 1,000 candidate pairs is computed. A third pair of corresponding epipolar line candidates, $(l_3,l_3')$, is selected based on this distance:
\begin{align}
(l_3,l_3')={\arg\min}\underset{(l_i,l_i')\in \{candidates\}\setminus \{(l_1,l_1'),(l_2,l_2')\}}{d(l_i,e)+d(l_i',e')}
\end{align}
The homography $H$ between the epipolar pencils is calculated by the homography DLT algorithm  \cite{hartleyMVG}, using the 3 proposed pairs of corresponding epipolar lines.

\subsection{Consistency Measure of Proposed Homography}
\label{section:Hconsistency}

Given the homography $H$, a consistency measure with all epipolar line candidates is calculated. This is done for each corresponding candidate pair $(l,l')$ by comparing the similarity between $l'$ and $\tilde{l'}= H l$. A perfect consistency should give $l' \cong \tilde{l'}$.

\begin{figure}
\centering
	\includegraphics[width=0.32\linewidth]{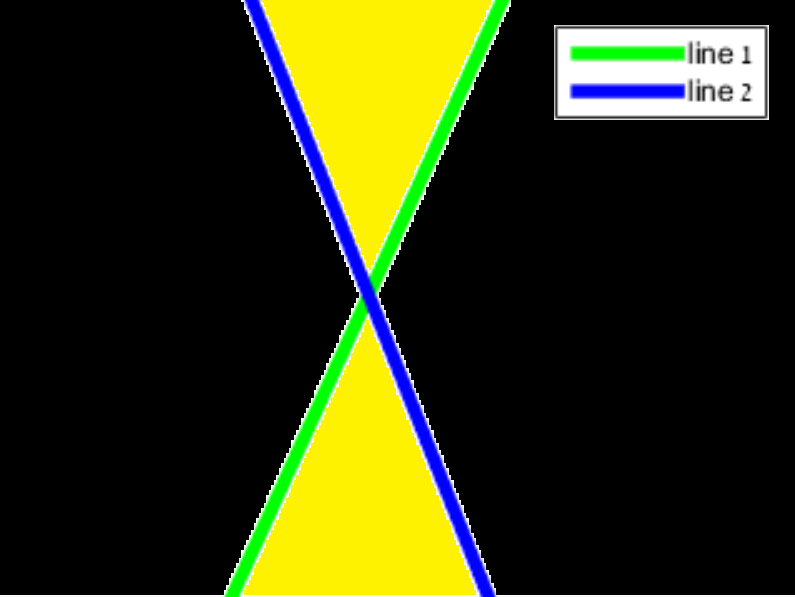}
\caption{Illustration of our distance measure between two lines. The distance measure between line 1 ({\it green line}) and line 2 ({\it blue line}) is the image area enclosed between the two lines ({\it yellow area}). }.
\label{fig:distance_area}
\end{figure}

Each candidate line $l$ in  $A$ is transformed to  $B$ using the homography $H$ giving $\tilde{l'}= H l$. We measure the similarity in  $B$ between $l'$ and $\tilde{l'}$ as the area between the lines (illustrated in Fig~\ref{fig:distance_area}).

The candidate pair $(l, l')$ is considered an inlier relative to the homography $H$ if the area 
 between  $l'$ and $\tilde{l'}$ is smaller than a predefined threshold. In the experiments in Sec.~\ref{section:experiments} this threshold was taken to be 3 pixels times the width of the image. The consistency score of $H$ is the number of inliers among all candidate lines.

\section{Experiments}
\label{section:experiments}

We tested our method on both synthetic and real video sequences. We created two synthetic datasets: \emph{cubes} and \emph{thin cubes},  using the Model Renderer that was developed by Assif and Hassner and was used in \cite{hassner2013viewing}. Each Cube dataset contains multiple views of a synthetic scene with flying cubes, while \emph{thin cubes} dataset has smaller cubes. Background subtraction is done automatically using the tool. 
As a real dataset we used PETS2009 \cite{pets2009}, using \cite{cucchiara2003detecting} for background subtraction. All datasets have $800$ synchronized video frames, recorded by multiple cameras.

These  datasets cannot be calibrated using matching of image features (e.g. SIFT), since there is no  background. The datasets cannot be calibrated by \cite{sinha2010camera,ben2015camera} since they have multiple objects, causing problems with the convex hull. The cubes datasets can not be calibrated by \cite{stein1999tracking,krahnstoever2005bayesian,lv2006camera,chen2007accurate} since the assumption of planar motion does not hold.    


The approach described in Sec.~\ref{section:method} was applied to each pair of cameras from each dataset. Initial lines were generated by uniformly sampling two points on the border of Cameras A and B, where Every two points sampled define a line passing through the image. 


\subsection{Consistency of the epipolar line pairs candidates}
Using the algorithm from Sec.~\ref{section:method}, 1,000 pairs of corresponding epipolar lines candidates were generated.

The simplest distance measure between a candidate line to a true epipolar line
is the distance of the candidate line from the true epipole. But this distance does not take into account the distance of the epipole from the image, and is inappropriate for epipoles that are far from the image. 
Instead, we measure the image area between the candidate line and a true epipolar line: the epipolar line going through the midpoint of the candidate line. This distance measure is illustrated in Fig~\ref{fig:distance_area}.

If this area is smaller than 3 times image length then the candidate line is considered a true positive. We call a pair of corresponding epipolar lines true if both lines are true. For each pair of cameras in each dataset we measured the true positives rate from all the 1,000 candidates, after removing the lines of motion (Sec~\ref{section:motionLines}). The average rate of true positives from each dataset is as follows: \emph{thin cubes}: $67.8\%$, \emph{cubes}: $71.67\%$ and  \emph{pets2009}: $37.81\%$.



\subsection{Multiple Objects Moving Same Straight Path}
\label{section:motionLines}

\begin{figure}
\centering
\includegraphics[width=0.32\linewidth]{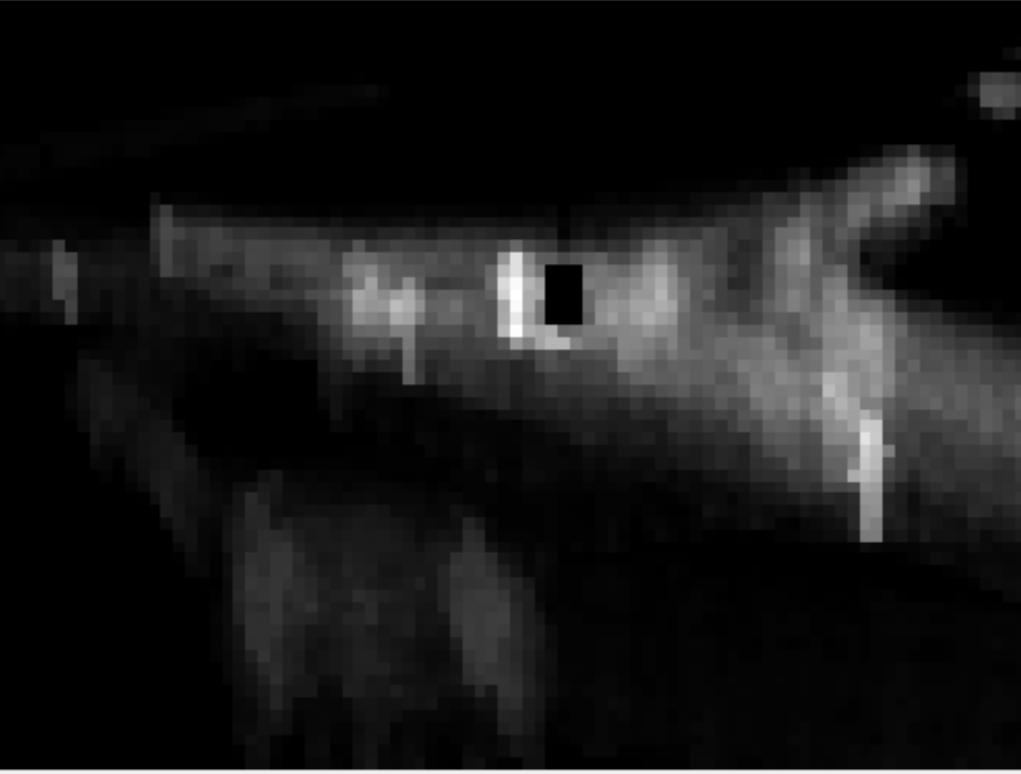}
\includegraphics[width=0.32\linewidth]{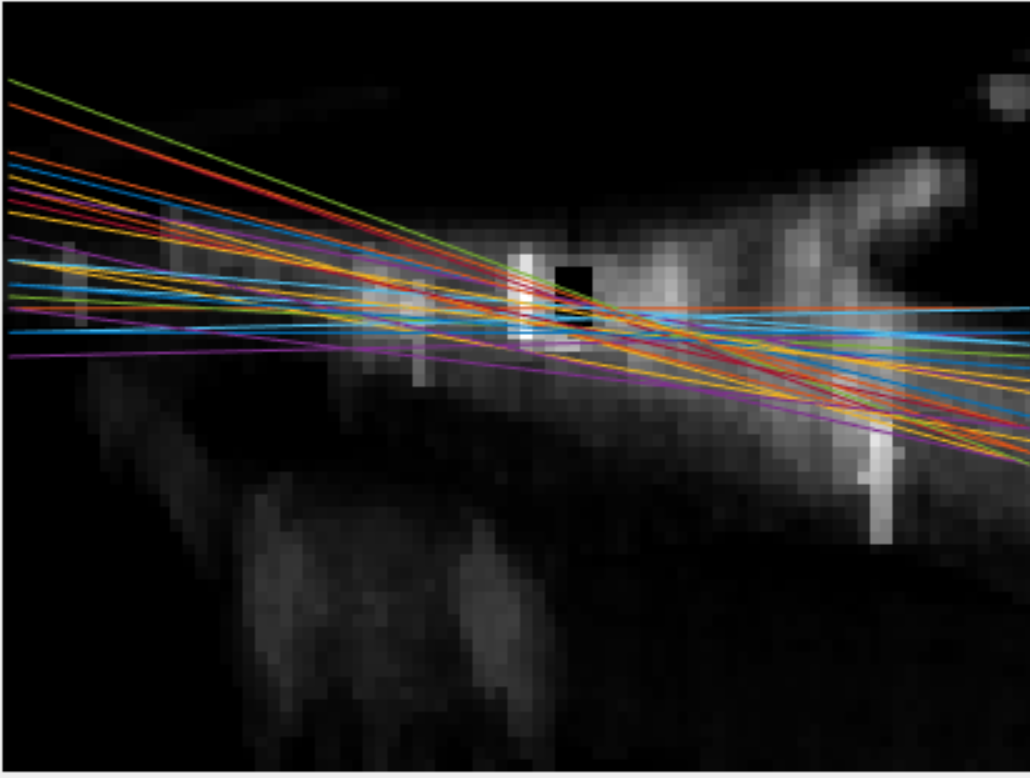}\\
(a) ~~~~~~~~~~~~~~~~~~~~~~~~~~~~~~ (b)
\caption{Detection of motion along lines for Camera 1 of Pets2009 dataset.
(a) Heat map obtained by summing the segmented frames over time.
(b) Line detection on the heat map gives motion along lines. Lines close to these lines will not be used as epipolar lines.}
\label{fig:motionLines}
\end{figure}

In some cases, e.g. busy roads, many objects in the scene may move in the same straight line. The projection of those lines on the video frames are straight lines. This may result in a high  correspondence between two non epipolar lines, as both will have similar Motion Barcodes. To overcome this problem, we create a motion heat map by summing all binary motion frames of each video to a single image. Candidate epipolar lines pairs, where both substantially overlap lines with heavy traffic in their corresponding images, are removed from consideration. See Fig~\ref{fig:motionLines}.

\subsection{Finding Fundamental Matrices}

After completing the procedure in Sec.~\ref{section:method}, we have 1,000 candidates for corresponding epipolar lines for each pair of cameras. Erroneous candidate lines, created by multiple objects moving in straight lines, are filtered out using the procedure in Sec.~\ref{section:motionLines}.

Given the candidates for corresponding epipolar lines, we perform up to  10,000 RANSAC
iterations to compute the fundamental matrix according to Sec.~\ref{section:ransac}. The quality of a generated fundamental matrix is determined by the number of inliers among all candidate pairs (See Sec.~\ref{section:Hconsistency}). We used the inlier threshold of $3$ times the length of the image for all the datasets.


We checked our method on the 3 datasets which contains 37 pairs of cameras. Except for one camera in Pets2009 dataset, all fundamental matrices were found with high accuracy.  One fundamental matrix for pair of cameras in Pets2009 could not be reproduced using our method. The reason is that all people in the scene are moving along one straight line which happens to be  projected to the corresponding epipolar lines in the images of the cameras. Although we get a high correlation between lines that are close to one of the epipolar lines, since the objects barely cross the epipolar lines, there is no pencil of corresponding true epipolar lines in the image, essentially there is  only one epipolar line pair which doesn't allow finding the true fundamental matrix.

For each resulting $F$ we checked its accuracy compared to the ground truth $F_{truth}$.
The accuracy was measured using Symmetric Epipolar Distance \cite{hartleyMVG}. By generating ground truth corresponding points using  $F_{truth}$, the Symmetric Epipolar error of  the resulting $F$, is measured on those points. Table ~\ref{table:errors} shows the results for the datasets that were tested. The table shows for each dataset, the average symmetric epipolar distance of all camera pairs, and how many camera pairs converged.

\begin{table} 
\begin{center}
\caption{Average symmetric epipolar distances for each dataset}
\label{table:errors}
\begin{tabular}{lcccl}
\hline\noalign{\smallskip}
Dataset &  Average error &  Number of \\
& & good pairs \\
\noalign{\smallskip}
\hline
\noalign{\smallskip}
Cubes  & 0.31 & $10/10$ &  \begin{minipage}{.12\textwidth}
      \includegraphics[width=\linewidth]{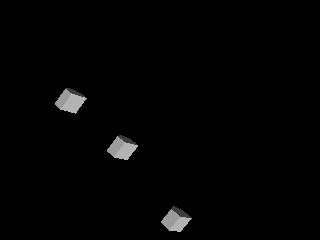}
    \end{minipage} \\\\
Thin Cubes & 0.79 & $21/21$ &  \begin{minipage}{.12\textwidth}
      \includegraphics[width=\linewidth]{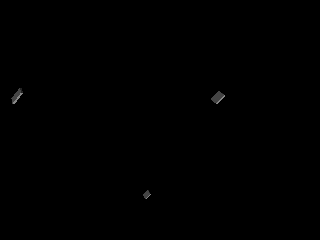}
    \end{minipage} \\\\
Pets 2009 & 1.69 & $5/6$ &  \begin{minipage}{.12\textwidth}
      \includegraphics[width=\linewidth]{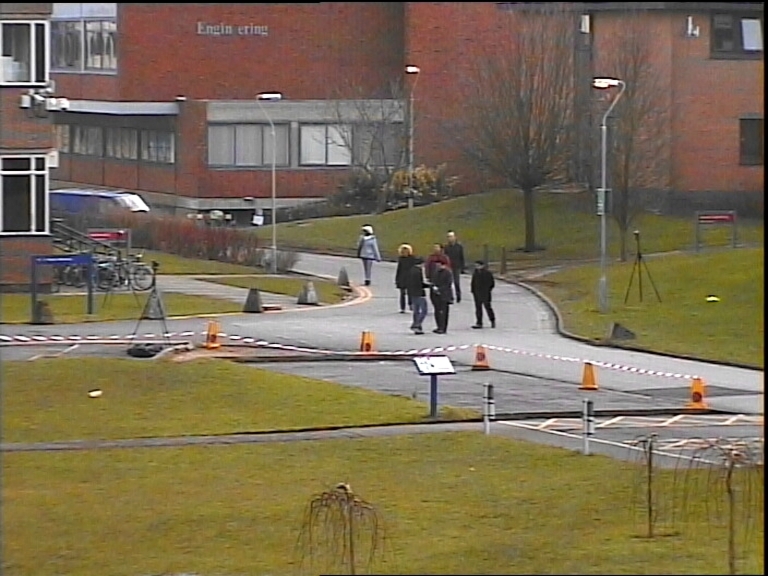}
    \end{minipage} \\\\
\hline
\end{tabular}
\end{center}
\end{table}


\section{Conclusions}

A method has been presented to calibrate two cameras having a very different viewpoints. The method has the following steps:
\begin{itemize}
\item Given two pairs of corresponding points, they are used to efficiently find candidates pairs of corresponding epipolar lines.
\item Using a RANSAC process, three corresponding pairs of epipolar lines are selected, and the fundamental matrix is computed.
\item A method to evaluate the quality of the fundamental matrix has also been proposed.
\end{itemize}

The proposed method is very accurate.

This method can be applied to cases where other methods fail, such as two cameras with very different viewpoints observing a scene with multiple moving objects. (i) Point matching will fail as the appearance can be very different from very different viewpoints. (ii) Silhouette methods will work on very different viewpoints, but only if they include a single object. 

~\\
\noindent
{\bf Acknowledgment.} This research was supported by Google, by Intel ICRI-CI, by DFG, and by the Israel Science Foundation.

\bibliographystyle{splncs}
\bibliography{egbib}
\end{document}